# On the referential capacity of language models: An internalist rejoinder to Mandelkern & Linzen

Giosuè Baggio[1], Elliot Murphy[2,3]

[1]Language Acquisition and Language Processing Lab, Department of Language and Literature, Norwegian University of Science and Technology, Trondheim, Norway
[2]Vivian L. Smith Department of Neurosurgery, University of Texas Health Science Center at Houston, TX, USA
[3]Texas Institute for Restorative Neurotechnologies, University of Texas Health Science Center at Houston, TX, USA

## 1. Summary of Mandelkern & Linzen (2024)

In a recent paper, Mandelkern & Linzen (2024) — henceforth M&L — address the question of whether language models' (LMs) words refer. Their argument draws from the externalist tradition in philosophical semantics, which views reference as the capacity of words to "achieve 'word-to-world' connections". In the externalist framework, causally uninterrupted chains of usage, tracing every occurrence of a name back to its bearer, guarantee that, for example, 'Peano' refers to the individual Peano (Kripke 1980). This account is externalist both because words pick out referents 'out there' in the world, and because what determines reference are coordinated linguistic actions by members of a community, and not individual mental states. The "central question to ask", for M&L, is whether LMs too belong to human linguistic communities, such that words by LMs may also trace back causally to their bearers. Their answer is a cautious "yes": inputs to LMs are linguistic "forms with particular histories of referential use"; "those histories ground the referents of those forms"; any occurrence of 'Peano' in LM outputs is as causally connected to the individual Peano as any other occurrence of the same proper name in human speech or text; therefore, occurrences of 'Peano' in LM outputs refer to Peano.

In this commentary, we first qualify M&L's claim as applying to a narrow class of natural language expressions. Thus qualified, their claim is valid, and we emphasise an additional motivation for that in Section 2. Next, we discuss the actual scope of their claim, and we suggest that the way they formulate it may lead to unwarranted generalisations about reference in LMs. Our critique is likewise applicable to other externalist accounts of LMs (e.g., Lederman & Mahowald 2024; Mollo & Millière 2023). Lastly, we conclude with a comment on the status of LMs as members of human linguistic communities.

## 2. Machines under public scrutiny: Why reference matters

The question of whether artificially generated words refer is a significant one, not least because, depending on our answers, we may or may not be able to claim that LMs' outputs are true or false, or carry true or false presuppositions. If a LM were to write a biography of Giuseppe Peano containing both true and false elements, we would not be able to say that what is true is true and what is false is false, if we took the proper name 'Peano' in the model's output *not* to refer to *any* individual in the world. Denying that machine speech and text *can* refer, or at least *be about* individuals and states of affairs, would undermine the all-important public enterprise of evaluating LM's outputs for truth and plausibility. In this sense, we suspect that recent arguments to the effect that, as a matter of principle, LMs cannot recover meaning from training data and infuse meaning into generated outputs (e.g., Bender & Koller 2020) could end up shifting the burden of truth onto human interpreters to an extent that may be problematic. If the string 'Peano taught in Bologna' does not mean or refer to anything 'for' a model that generates it, then it would be false only in virtue of interpretations it receives by human users: machine language would only be evaluable in an arena of competing interpreters and interpretations, not against the constraints of a language. One may reply that, minimally, we ought to interpret LMs as we would any human member of a linguistic community: 'Peano taught in Bologna' is false when it is produced by human speakers, and therefore must be false when generated by LMs too. This position is attractive, but needs to be qualified. We will outline such a qualification in Section 5, but before that we will examine M&L's arguments in greater detail.

## 3. "Words"? What words?

M&L's focus is on classical externalist examples: proper names and so-called natural kind terms, such as 'Peano', 'water', etc. There is no mention of other types of expressions in their article. This would not necessarily be a problem, if it did not also invite the unwarranted generalisation that reference functions in the same way for all or most 'words' in the language. We could not find a statement by M&L that restricts the scope of their argument and conclusions to only those classes of expressions to which broadly causal-historical theories of reference apply (whether they also succeed, is a different issue; see below).

There are causal histories of usage for all words in a language, and there is a *variety* of *ultimate anchors* for those histories. Not all such anchors are of the ontological types that externalism favours. To lean on some classical textbook examples, what would 'democracy', 'love', 'mathematics', 'English', etc. trace back to, if anything? Not to entities and events in the world, but rather *concepts* that organise information from cognitive and bodily states, (core) knowledge, actions, and so forth. Or consider expressions that require a *situated speaker* for felicitous production and interpretation, primarily indexicals, pronouns, and demonstratives. What would 'now' and 'here' refer to in LMs' outputs: where is a LM, such that 'here' may be interpreted as denoting its current location in space (Murphy 2016)? Personal pronouns are often only interpretable relative to hypotheses about the speaker's specific referential intentions: who are 'us' and 'them' in machine outputs, when LMs cannot form referential intentions? M&L touch upon the question of whether current LMs could "learn to model communicative intention in order to accurately predict upcoming words", yet this is distinct from the ability to *form* communicative intentions of one's own (Sperber & Wilson 2024). One could grant machines the capacity to generate and maintain 'simulacra' that would appear to have communicative intentions and other mental states, to the extent that LMs can "convincingly play the role of a character that does" (Shanahan et al. 2023). That could reduce the risk of anthropomorphising artificial dialogue systems, but it does not automatically endow LMs' outputs with viable referential properties in context: e.g., neither the LM nor the simulacra it supports appear to occupy the sorts of spaces that would make uses of 'I', 'here', etc. pick out contextually appropriate referents. Externalist logic then applies to an important but narrow class of expressions. In many cases, reference minimally requires a syntax or grammar interfacing with conceptual-semantic, pragmatic, and general cognitive systems (Murphy et al. 2024), and language users with communicative intentions (Scott-Phillips 2014), situated in social and physical space and time (Hinzen 2016).

## 4. Deeper problems with externalism

As noted above, M&L observe that LM inputs include "strings of symbols with certain natural histories which connect them to their referents". This concept of 'natural histories' cannot be subject to much interrogation given its artificial nature, particularly in the context of machine language, as there appears to be nothing natural, nor historical, about linear strings of tokens, as considered by current tokenisation algorithms. Tokens do not necessarily correspond to the kinds of expressions or parts of expressions that can have causal histories, such as characters, morphemes, and 'words' (Murphy 2024): e.g., the GPT-3.5 and GPT-4 tokeniser breaks up the word 'motionlessness' into 14 characters and 2 tokens: 'motion' and 'lessness'. But even ignoring the 'unnaturalness' of LMs' lexical (non)compositions, deeper problems arise with the general externalist thesis, endorsed by M&L, that "what grounds reference is the *natural histories* of words: the causal-historical links between a speech community's use of a word and the word's referent". This formulation has been unpacked and, we believe, thoroughly undermined in the literature (among others, see Chomsky 2000; Jackendoff 2002; Pietroski 2017, 2018; Baggio 2018; Murphy 2023).[1]

Speech communities are appropriately underspecified notions for social scientists, but not for those concerned with a biologically grounded theory of language. What delimitations can we place on the notion of 'community', that would help with explaining the semantic properties of a language, including what words refer to? Coherent and coordinated linguistic actions by members of a community are *constrained* by how our cognitive apparatus carves up and conceptualises reality. Even if *first-order externalism* was true (the reference of some linguistic expressions used by a speaker S is partly determined by factors independent of S's mental states; Cohnitz & Haukioja 2013), *meta-internalism* would still be applicable (how an expression E in some utterance U by S refers, and which theory of reference is true of E, are determined by S's mental states at the time of U). In fact, its negation has been convincingly argued to be "an implausible view about semantics" (Cohnitz & Haukioja 2013). Facts about the mind-brain must be invoked to explain *how* words refer: not *what they refer to*, but *what determines what they refer to* (Block 1987; Carey 2009; Baggio 2018).

M&L claim that machines may achieve referential capacities by "standing in the right kind of natural history to the referent". They say that if a friend, Luke, sends us the text message 'Peano proved that arithmetic is incomplete', "his text *says something false*, namely something *about Peano*: that he proved incompleteness. So Luke's use of the word 'Peano' refers to Peano." Of course, we agree with this, even though we want to frame the observation differently. Under the assumption that 'Peano' refers to Peano, in a language or an idiolect where that is the case, that sentence is false. That is, in this context, Luke's text says something false according to our internal models of the world and how our language capacity interfaces with them. We get the intuitive feeling that Luke is — whether he wishes to or not, and whether what he says is true or not — *referring to* Peano: our own generative inferences are *about* Peano. Different cognitive models of the world, even possibly two idiolects or I-languages, can conflict with respect to what words denote in conceptual or (core) knowledge spaces. People can communicate not because they 'share a language' in which words are (causally) connected to their referents, but rather because individual I-languages overlap to an extent sufficient to make communication possible (McGilvray 1998; Chomsky 2000; Segal 2000; Pietroski 2003; Baggio 2018).

---

[1] Relatedly, M&L review Putnam's "striking" Twin Earth thought experiment. Several critiques have challenged Putnam's argument (e.g., Chomsky 2000; Pietroski 2017, 2021; Murphy 2023).

M&L discuss cases where machines have been exclusively trained on the kind of orthographic data that Luke may have received about Peano and other historical figures. They write: "*The inputs to LMs are not just forms, but forms with particular histories of referential use*. And those histories ground the referents of those forms, whether or not you know them or have any kind of access to them." They then ask what the philosophical implications of this are. Yet, they do not ask a more relevant question: would a LM trained only on photographs and 'tokens' of ant trails accidentally spelling out English sentences yield any philosophical import, even if it achieved comparable accuracy and fluency to a model trained on Wikipedia, and regardless of whether it generated true or false sentences about historical figures? We consider the answer fairly obvious.

Moreover, externalist theories of reference require additional stipulations *in their favour,* where words can provide humans and machines with *extensions* that have to be learned and tested continuously during processing. Extensions can be fairly straightforwardly stated for some expressions, like proper names, but are much harder to define for most others, including natural kind terms. An alternative framework, provided by theoretical linguistics and internalist philosophy of language (e.g., Jackendoff 2002; van Lambalgen & Hamm 2004; Pietroski 2018, 2021; Baggio 2018; Pustejovsky & Batiukova 2019; Murphy 2023), assumes that lexical items (words) and their compositions are *algorithms* that can instruct cognitive brain systems for thought, planning, reasoning, and so forth.[2] Conversely, when we lexicalise a concept, we may subtly *change* it or *reformat* it by placing it in a domain-specific intensional space (Pietroski 2018).

When we utter a sentence like (1), we are not necessarily presupposing any connections to states of the external world, but to states within our own *models* of the world, however one wants to construe such models:

(1) The previous emperor of Kansas is about to announce their new mixtape.

Language provides 'truth-indications', rather than truth-evaluations. Truth is intimately and intrinsically a *syntactic* or *grammatical* phenomenon (Hinzen & Sheehan 2013): only structures of a certain level of syntactic complexity may be readily judged to be true or false (e.g., complex Tense Phrases, but not basic Noun Phrases in isolation, which would require pragmatic enrichment).

Just as how there cannot be an entity that is simultaneously concrete and abstract (Gotham 2016; Murphy 2023), there also cannot be events that exhibit contradictory features.[3] Consider an event where John and James are hunting each other. We may say that the sentences in (2) accurately describe the event:

(2) a. John chased James athletically but not skillfully.
    b. James chased John unathletically but skillfully.

However, no event could jointly host all of these features: an action cannot be skillful and unskillful, a person cannot be athletic and unathletic at the same time. This has less to do with properties of the 'external world', as assumed by most externalists, than with peculiarities of human cognition (Pietroski 2005). What kind of 'natural history' is required by LM inputs to achieve successful reference to an extensional treatment of this event? What may be the ultimate anchors, the endpoints of causal histories, for 'athletical' and 'skillfull'? These and other adjective types, along with the examples from Section 3 and others, resist an externalist analysis. Externalism lacks general cognitive applicability, putting aside even more fundamental considerations. Externalist accounts of machine behaviour, which seek commensurability with human cognition, will likewise be stymied. However, current LMs, unlike human minds and brains, cannot be understood (meta)internalistically either (Bever et al. 2023), lacking as they do the kinds of cognitive systems required to constrain and infuse key aspects of meaning into a sufficiently broad range of expressions.

---

[2] This is different from Ludlow's (2003) distinction between types of reference: lexical items can 'refer' to other cognitive systems, or they could refer to actual properties of the world.
[3] The work of Gotham (2016, 2022) here provides a potent series of arguments in our favour.

Most of our objections apply throughout M&L's paper. For example, they consider the case of ants unintentionally carving out a long path that spells out 'Peano proved that arithmetic is incomplete'. In contrast, as above, we receive a text from Luke saying the same thing. M&L remark that "Luke's words mean something definite on their own (regardless of whether you or anyone else interprets them): namely, that Peano proved that arithmetic is incomplete. What Luke said is false: it was Gödel who proved incompleteness. But Luke said something, whereas the ants didn't say anything at all."

The claim that Luke's words "mean something on their own" confuses the external 'sense data' of orthographic patterns on a screen, or another material support, with "meaningful words", the *intensions* (internalised algorithms) we trigger *from* such data (Baggio 2018; Murphy 2024).[4] Luke said something true only *if there was an utterance*, and only if this was his communicative intention: the 'words' would not do any referring, for the same reason that orthographic information 'generated' by the ants does not do any referring. Relatedly, 'lion' would not refer to anything in isolation (Hinzen 2016). Only when placed in a particular grammatical configuration (e.g., 'that lion') can *we* refer *with it*.

The premise of M&L's main argument is set on a problematic foundation. They write: "We are interested in the question of whether the outputs of LMs' are more like the ants' patterns or like Luke's text: do they merely *resemble* meaningful sentences, or are they in fact meaningful sentences?" But what is the difference between merely resembling a meaningful sentence and being one? Luke's message *also* 'resembles' a meaningful sentence: how do we know that our phone is not infected by a virus or bug, and is genuinely showing the results of a human's communicative action? *Both* the ant's trail and Luke's text provide the exact same type of data to our language faculty. Naturally, in the case of Luke, we suspect that there was a referential intent, and so pragmatics kicks into action (Sperber & Wilson 2024). But the difference is in our internal states, that are triggered by or accompany (similar) data (Hagoort 2023).

---

[4] One of us (E.M.) once asked a literary theorist where they stood on the question of meaning. Where does it reside? In the text? Or in the mind of the reader? Is it socially constructed? Is it provided purely by 'context'? The magnificent reply given after a thoughtful pause ("I think the text has some work to do") gestures towards a widespread category mistake, consisting in seeing philosophical implications for reference in the surface forms of linguistic data.

As noted in Section 3, language models patently lack the capacity to form communicative intent, and so any such orthographic output from them could only be truth-evaluable in relation to our own language capacity's judgments. M&L claim that "nearly all agree" that meaning requires reference, but this is true only if we assume that the language faculty 'indexes' words to conceptual entries for activation, composition, and inference. We do not have to stipulate that words enjoy some kind of supra-natural 'direct' relationship to entities in the external world, nor do we have to assume that they do so in virtue of their existence in some extra-mental realm. Neither language models themselves can refer, and nor can 'their words' refer. We consider these observations to be fairly trivial from the standpoint of contemporary psycholinguistics and theoretical linguistics, and we see little in the way of profound philosophical import here. The question asked by M&L ("whether the word 'Peano' output by the LM refers to Peano, or rather doesn't refer at all") rests on an invalid notion of reference (Ludlow 2003; Murphy 2014). As was argued in the 1950s, words do not refer; people (with intentions, world models, etc.) do (Strawson 1950). M&L speak of "referential contact with the external world", which we suspect to be unachievable in general by either humans or machines. With the suggestion that machines can deploy words like 'Peano' to refer to the external world, "there is a failure of that feeling for reality which ought to be preserved even in the most abstract studies" (Russell 1905).

## 5. Crawl out through the fallout: Machines in linguistic communities

In the *Fallout* universe, the robot assistant Codsworth's voice was taken from a recording of a human being long since deceased. It animates the wasteland with echoes of this voice, providing the illusion of it being part of a linguistic community, when in fact it is only a stochastic reassembly of sampled speech. The inclusion of LMs as *bona fide* members of linguistic communities is not unlike Codsworth's ability to bring up encyclopedia entries in its memory and recite historical facts: both are based upon the artificial agents' capacity to 'role play', or simulate characters that appear to have the mental states that users of these systems would impute to humans showing comparable behaviours. The question is what counts as a 'member of a linguistic community': the LM, the simulacra that the LM supports, both, or neither? LMs as such would be highly atypical members of linguistic communities. They lack the ability to identify themselves in social and physical space (e.g., through indexicals) and to form intentions. Extending the notion of 'linguistic community member' to include LMs can be a valuable conceptual engineering exercise, but it does not address the issue whether LMs have the required status in *our* communities, such that some of their 'words' can refer in the ways envisaged by externalist theories.

Recall that the pressing question is whether and how expressions, such as proper names and natural kind terms, 'refer' in LMs outputs, such that we can critically assess those outputs for truth and plausibility. This is urgent also in connection with the growing use of language technologies in research (e.g., in literature summarisation, or hypothesis generation), where assessing just how 'epistemically trustworthy' LMs' outputs are is key (Messeri & Crockett 2024). Here, two ideas might be worth considering. The first is that it is *simulacra* that would count as members of human linguistic communities. It is these virtual characters, and not LMs, that we interact with through language, and it is the quality of their outputs which we have to try to evaluate. If an AI summarises scientific articles about electrons, we want to take 'electron' in its summaries as 'referring to' electrons (a postulate in an explanatory scientific theory). The same may apply to outputs of other tasks, such as hypothesis generation, and to some other expressions (e.g., proper names, but not indexicals). Only then would we be in a position to argue that a summary is not only good or bad qua summary, but also contains true or false, plausible or implausible statements about electrons. The *virtual* agent thus makes statements about the *real* world: this makes it a more viable member of a linguistic community than, say, some character in a science fiction book who talks about electrons, but whose claims we could only assess in the fictional world of the novel, where electrons might have properties they lack in the actual world or vice versa. However, from our observations in Section 4 follow general qualms about reference, which would make it difficult for us to accept the argument that even *some* of virtual agents' words can *refer*, in the sense of "achieve 'word-to-world' connections": LMs, *a fortiori* the virtual agents they support, simply lack the mental structures that mediate between language and the world, and that are necessary for people to be able to refer to aspects of the world.

One way out is to argue that LM's virtual agents are not able to refer, but their outputs are *about* the same subject matters those claims would be about, were they generated by humans. *Aboutness* is more directly applicable to LMs than reference: LMs' performance in topic modelling tasks is easily explained by positing internal states that are less problematic philosophically, and that do not invite unwarranted comparisons to human cognition. In the example above, the AI's text will be *about* electrons. This does not imply that the AI can refer to electrons, only that the strings it generates *concern* electrons, for which human language has a rigid designator in the term 'electron' (Hawke 2018).

## Acknowledgements

We thank Matthew Mandelkern and Tal Linzen for their comments on this paper.